\documentclass{article}
\usepackage{iclr2022_workshop,times}

\iclrfinalcopy

\usepackage{amsmath,amsfonts,bm}

\def\eqref#1{equation~\ref{#1}}
\def\1{\bm{1}}

\DeclareMathAlphabet{\mathsfit}{\encodingdefault}{\sfdefault}{m}{sl}
\SetMathAlphabet{\mathsfit}{bold}{\encodingdefault}{\sfdefault}{bx}{n}

\usepackage[hidelinks,breaklinks=True]{hyperref}
\usepackage{url}

\usepackage{breakurl}

\title{Sparsifying the Update Step in Graph Neural Networks}

\author{Johannes F. Lutzeyer\thanks{Both authors contributed equally to this research.} , ~Changmin Wu\footnotemark[1]~ \& ~Michalis Vazirgiannis\\
DaSciM, Laboratoire d'informatique, École Polytechnique, Institute Polytechnique de Paris, France\\
\texttt{\{johannes.lutzeyer,changmin.wu\}@polytechnique.edu}\\
\texttt{mvazirg@lix.polytechnique.fr}
}
\usepackage{tabularx}
\usepackage{multirow}
\usepackage{amssymb, amsmath, amsthm} % blackboard math symbols
\usepackage{bm}
\usepackage{graphicx}

\theoremstyle{definition}

\newcommand{\expander}{{{\itshape Expander}}}
\newcommand{\activation}{{{\itshape Activation-Only} }}

\begin{document}

\maketitle

\begin{abstract}
Message-Passing Neural Networks (MPNNs), the most prominent Graph Neural Network (GNN) framework, celebrate much success in the analysis of graph-structured data. Concurrently, the sparsification of Neural Network models attracts a great amount of academic and industrial interest. In this paper we conduct a structured, empirical study of the effect of sparsification on the trainable part of MPNNs known as the Update step. To this end, we design a series of models to successively sparsify the linear transform in the Update step. Specifically, we propose the ExpanderGNN model with a tuneable sparsification rate and the Activation-Only GNN, which has no linear transform in the Update step. In agreement with a growing trend in the literature the sparsification paradigm is changed by initialising sparse neural network architectures rather than expensively sparsifying already trained architectures. Our novel benchmark models enable a better understanding of the influence of the Update step on model performance and outperform existing simplified benchmark models such as the Simple Graph Convolution. The ExpanderGNNs, and in some cases the Activation-Only models, achieve performance on par with their vanilla counterparts on several downstream tasks, while containing significantly fewer trainable parameters. Our code is publicly available at: \url{https://github.com/ChangminWu/ExpanderGNN}.
\end{abstract}

\section{Introduction}\label{sec:intro}
In recent years we have witnessed the blossom of Graph Neural Networks (GNNs). They have become the standard tools for analysing and learning graph-structured data \citep{wu2020comprehensive} and have demonstrated convincing performance in various application areas, including chemistry \citep{duvenaud2015convolutional}, social networks \citep{monti2019fake}, natural language processing \citep{yao2019graph} and neural science \citep{griffa2017transient}.

Among various GNN models, Message-Passing Neural Networks (MPNNs, \citet{gilmer2017neural}) and their variants are considered to be the dominating class. In MPNNs, the learning procedure can be separated into three major steps: 
{\itshape Aggregation}, {\itshape Update} and {\itshape Readout}, where {\itshape Aggregation} and {\itshape Update} are repeated iteratively so that each node's representation is updated recursively based on the transformed information aggregated over its neighbourhood. 
There is thus a division of labour between the {\itshape Aggregation} and the {\itshape Update} step, where the {\itshape Aggregation} utilises local graph structure, while the {\itshape Update} step is only applied to single node representations at a time independent of the local graph structure. From this a natural question arises: \textit{What is the impact of the graph-agnostic {\itshape Update} step on the performance of GNNs? }
Since the {\itshape Update} step is the main source of model parameters in MPNNs, understanding its impact is fundamental in the design of parsimonious GNNs.

% In an attempt to answer this question we develop a series of models which successively remove trainable parameters from the {\itshape Update} step and thus allow us to assess the impact of the {\itshape Update} step in several MPNN frameworks used for node and graph classification as well as graph regression. We furthermore find that our sparisified models often perform on par with the standard MPNNs and 

% \citet{wu2019simplifying} first challenged the role of the {\itshape Update} step by proposing a Simple Graph Convolution (SGC) model
% % simplified graph convolutional network (SGC) 
% where they removed the non-linearities in the {\itshape Update} steps and collapsed the consecutive linear transforms into a single transform. Their experiments showed, surprisingly, that in some instances the {\itshape Update} step of Graph Convolutional Network (GCN, \citet{kipf2016semi}) can be left out completely without the models’ accuracy decreasing. 

In this paper we empirically analyse the impact of the {\itshape Update} step and its sparsification in a systematic way. To this end,
we propose two nested model classes, where the {\itshape Update} step is successively sparsified. In the first model class which we refer to as \expander GNN, the linear transform layers of the {\itshape Update} step are sparsified; while in the second model class, the linear transform layers are removed and only the activation functions remain in the model. We name the second model \activation GNN and it contrasts the Simple Graph Convolution model (SGC, \citeauthor{wu2019simplifying}, \citeyear{{wu2019simplifying}}) where the activation functions where removed to merge the linear layers.

% Inspired by the recent advances in the literature of sparse Convolutional Neural Network (CNN) architectures  \citep{prabhu2018deep}, we propose to utilise a random sparsification scheme, which is motivated by the study of expander graphs (hence the model's name). 
% % (hence the model name:  \expander GNN). 
% Here the sparsification is performed at intialisation and accordingly saves the cost of more traditional methods, which often iteratively prune connections during training.   
% % Guided by positive graph theoretic properties, %graph theory,
% % it optimises sparse network architectures at initialisation and accordingly saves the cost of traditional methods of iteratively pruning connections during training.   

Through a series of empirical assessments on different graph learning tasks (graph and node classification as well as graph regression), we demonstrate that the {\itshape Update} step can be heavily simplified without inhibiting performance or relevant model expressivity. Our findings partly agree with the work in \citet{wu2019simplifying}, in that dense {\itshape Update} steps in GNN are expensive and often ineffectual. In contrast to their proposition, we find that there are many instances in which leaving the {\itshape Update} step out completely significantly harms performance. In these instances our \activation model shows superior performance while matching the number of parameters and efficiency of the SGC.

\section{Related Work}\label{sec:rw}
In recent years the idea of {\itshape utilising expander graphs in the design of neural networks} is starting to be explored in the CNN literature. Most notably, \citet{prabhu2018deep} propose to replace linear fully connected layers in deep networks using an expander graph sampling mechanism and hence, propose a novel, well-performing CNN architecture they call X-nets. 
% The great innovation of this approach is that well-performing sparse neural network architectures are initialised rather than expensively calculated. 
% Furthermore, they are shown to compare favourably in training speed, accuracy and performance trade-offs to several other state-of-the-art architectures. 
\citet{McDonald2019} and \citet{Robinett2019} build on the X-net design and propose alternative expander sampling mechanisms to extend the simplistic design chosen in the X-nets.
% Independent of this literature branch, \citet{Bourely2017} explore 6 different mechanisms to randomly sample expander graph layers. 
% Across the literature the results based on expander graph layers are encouraging. 

The  \textit{Sparisification and Pruning of neural networks} is a very active research topic \citep{Hoefler2021, Blalock2020}. In particular, a wealth of algorithms sparsifying neural network architectures at initialisation has recently been proposed \citep{tanaka2020pruning, Wang2020, Lee2019}. While these algorithms make pruning decisions on a per-weight basis,  \citet{Frankle2021} find that these algorithms produce equivalent results to a per-layer choice of a fraction of weights to prune, as is directly done in our chosen sparsification scheme. 
All of these research efforts are pruning CNNs, typically the VGG and ResNet architectures. \textit{To the best of our knowledge, our work is the first investigating the potential of sparsifying the trainable parameters in GNNs.} %This allows us to observe that %conclusions drawn for CNNs do not directly carry over to GNNs. We observe that 
% in the majority of cases pruning 90\% of trainable weights at initialisation using our relatively simple sparsification scheme comes at no performance cost. %The presence of the \textit{Aggregation} step in the graph neural network architecture appears to  significantly impact the response of the model to pruning.   

Both \citet{wu2019simplifying} and \citet{Salha2019} observed that {\itshape simplifications in the Update step of the Graph Convolutional Network (GCN, \citeauthor{kipf2016semi}, \citeyear{kipf2016semi}) model} is a promising area of research. % \citep{wu2019simplifying,Salha2019}.
\citet{wu2019simplifying} proposed the SGC model, where %Simple Graph Convolution (SGC) model, where %as a simplification of the GCN architecture. Their
simplification is achieved by removing the non-linear activation functions from the GCN model. This removal allows them to merge all linear transformations in the {\itshape Update} steps into a single linear transformation without sacrificing expressive power. 
\citet{Salha2019} followed a similar rationale in their simplification of %found an agreeing results for 
the graph autoencoder and variational graph autoencoder models. 
% These works
% %The SGC and Linear Graph AE 
% have had an immediate impact on the literature featuring as benchmark models and object of study in many recent papers: The idea of omitting the {\itshape Update} step %served as %
% guided \citet{chen2020powerful} in the design of simplified models and has found successful application in various areas where model complexity needs to be reduced \citep{waradpande2020deep,he2020lightgcn} or very large graphs  %($\sim10^6$ nodes/edges) 
% need to be processed \citep{salha2020simple}.
In our work we aim to extend these efforts by providing %several%
more simplified benchmark models for GNNs without a specific focus on the GCN. %the established message passing neural networks. 

%%%%%%%%%%%%%%%%%%%%%%%%%%%%%%%%%%%%%%%%%%%%
%%%%%%############################%%%%%%%%%%
%%%%%%%%%%%%%%%%%%%%%%%%%%%%%%%%%%%%%%%%%%%%
\section{Investigating the Role of the Update step}\label{sec:model}
In this section, we present the general model structure of MPNNs and our two proposed model classes, where we sparsify or remove the linear transform layer in the {\itshape Update} step. 

We define graphs $\mathcal{G}=(\bm{A},\bm{X})$ in terms of their adjacency matrix $\bm{A}=[0,1]^{n\times n},$ which contains the information of the graph's node set $\mathbb{V},$ and the node features $\bm{X}\in\mathbb{R}^{n\times s}.$ 
The MPNN structure is a prominent paradigm for performing  machine learning tasks on graphs such as node or graph classification. 
The learning procedure of MPNNs can be divided into the following phases:
\paragraph{Initial (optional).} In this phase, the initial node features $\bm{X}$ are mapped from the feature space to a hidden space by a parameterised neural network $U^{(0)},$ usually a fully-connected linear layer;
 $$\bm{H}^{(1)} = U^{(0)}(\bm{X}) = \left(\bm{h}^{(1)}_{1}, \ldots, \bm{h}^{(1)}_{n}\right), $$
% $\bm{H}^{(1)} = U^{(0)}(\bm{X}) = (\bm{h}^{(1)}_{1}, \ldots, \bm{h}^{(1)}_{n}), $
where $\bm{h}^{(1)}_{i}$ denotes the hidden representation of node $i.$  %, which will be used as the initial point for later iterations.

\paragraph{Aggregation.} For each node, information is gathered from the node's neighbourhood, denoted $\mathcal{N}(i)$ for node $i.$ The gathered pieces of information are called ``messages'', denoted by $\bm{m}_{i}$. Formally, if  $f^{(l)}(\cdot)$ denotes the aggregation function at iteration $l,$ then
\begin{displaymath}
    \bm{m}_i^{(l)} = f^{(l)}\left(\left\{\bm{h}_j^{(l)} | j\in\mathcal{N}(i)\right\}\right).
\end{displaymath}
% $\bm{m}_i^{(l)} = f^{(l)}(\{\bm{h}_j^{(l)} | j\in\mathcal{N}(i)\}),$
% where $f^{(l)}(\cdot)$ is the aggregation function at iteration $l$. 
Due to the isotropic nature of graphs (arbitrary node labelling), this function needs to be permutation equi- or invariant. It also has to be differentiable so that the framework will be end-to-end trainable. 

\paragraph{Update.} The nodes then update their hidden representations based on their current representations and the received ``messages''. Let $U^{(l)}$ denote the neural network learning the update function at iteration $l.$ Then, for node $i,$ we have
\begin{displaymath}
\bm{h}_i^{(l+1)} = U^{(l)}\left(\bm{h}_i^{(l)}, \bm{m}_{i}^{(l)}\right).
\end{displaymath}
% $\bm{h}_i^{(l+1)} = U^{(l)}(\bm{h}_i^{(l)}, \bm{m}_{i}^{(l)}),$
% where $U^{(l)}$ is the update function at iteration $l$. 

\paragraph{Expander Update Step.} If we denote the two sets of computational units, i.e., neurons, being connected in a fully-connected layer of a neural network by  $\mathbb{S}_1$ and $\mathbb{S}_2,$ respectively. Then, following  \citet{prabhu2018deep}, we propose to sparsify this layer by uniform randomly sampling only $d$ connections for the computational units in the smaller of the two connected sets $\mathbb{S}_1$ and $\mathbb{S}_2.$
We refer to the \textit{density} of such an expander linear layer as the ratio of the number of sampled connections to the number of connections in the fully-connected layer, i.e.,  $d/\max(|\mathbb{S}_1|,|\mathbb{S}_2|).$ 
The theoretical computational cost of an expander linear layer is equal to $2nd \min(|\mathbb{S}_1|, |\mathbb{S}_2|)$ Floating Point Operations (FLOPs). % for $|\mathbb{S}_1|\leq|\mathbb{S}_2|$ and $2nd|\mathbb{S}_2|$ FLOPs if $|\mathbb{S}_1|\geq|\mathbb{S}_2|.$ 
The tunable parameter $d$ can therefore lead to significant computational savings as the computational cost of a fully connected linear layer equals $2n|\mathbb{S}_1||\mathbb{S}_2|$ FLOPs. 
When we replace all linear layers in the {\itshape Update} steps of a GNN with expander linear layers, we get the \expander GNN.

When compared to pruning algorithms which sparsify neural network layers by iteratively removing parameters according to certain metric during training, the expander sparsifiers have two advantages:
\begin{enumerate}
    \item %Good properties of {\itshape expander structures} allow consecutive linear layers to be highly connected when only a smaller number of edges is present. 
    The expander design assures that paths exist between consecutive layers, avoiding the risk of {\itshape layer-collapse} that is common in many pruning algorithms, where the algorithm prunes all parameters (weights) in one layer and cuts down the flow between input and output \citep{tanaka2020pruning}.      
    \item The expander sparsifier removes parameters at initialisation and keeps the sparsified structures fixed during training, which avoids the expensive computational cost stemming from adapting the neural network architecture during or after training and then retraining the network as is done in the majority of  pruning algorithms \citep{Frankle2019, Han2015}.   
\end{enumerate}
Theoretical motivation and implementation details of the Expander Update Step can be found in Appendix Section \ref{app:mot_imp_exgnn}.

\paragraph{Activation-Only Models.}
In the \activation GNN models, we propose to fully remove the linear transformation in the {\itshape Update} step such that 
each message-passing step is immediately followed by a pointwise activation function.
The resulting model can be seen as a natural extension of the \expander GNN, where the linear transformation of the {\itshape Update} step is completely forgone.

% A brief theoretical motivation and details about the implementation of our two proposed models can be found in Appendix Section \ref{app:mot_imp_exgnn}.

% \paragraph{Motivation of the Activation-Only Models.} 
In \citet{fern2019stability} it is argued that the non-linearity present in GNNs, in form of the activation functions, has the effect of frequency mixing in the sense that ``part of the energy associated with large eigenvalues'' is brought ``towards low eigenvalues where it can be discriminated by stable graph filters.'' 
% Hence, the SGC where the activation functions are removed is not able to capture the information stored in the high energy part of graph signals. 
The theoretical insight that activation functions help capture information stored in the high energy part of graph signals %into the role of activation functions in GNNs
is strong motivation to consider an alternative simplification to the one made in the SGC. In this alternative simplification, which we refer to as the \activation GNN models, we remove linear transformations instead of activation functions such that each message-passing step is immediately followed by a pointwise activation function.

\paragraph{Readout (optional).} After $L$ aggregation and update iterations, depending on the downstream tasks, the MPNN will either output node representations directly or generate a graph representation via a differentiable readout function,
$$
    \bm{g} = R\left(\left\{\bm{h}^{(L)}_i | i\in\mathbb{V}\right\}\right).
$$

\section{Experiments and Discussion} \label{sec:experiments}

In order to investigate the influence of the {\itshape Update} step in GNNs,  we now observe the performance of the proposed benchmark models on the tasks of graph classification, graph regression and node classification. In the Appendix Section \ref{sec:exp_general_setting} we provide an overview of our experimentation setup.   We have made our experimentation code publicly available online\footnote[1]{\url{https://github.com/ChangminWu/ExpanderGNN}}.%We make our anonymised experimentation code accessible online\footnote{\url{https://www.dropbox.com/sh/hzb1x8bv6o6t1o6/AAAkjANtZy_1nJ2jGqvAMkQ3a}}.

% At last, in Section~\ref{sec:sparsification}, we discuss the potential benefit that \expander models can bring. 

\begin{table*}[t]
\caption{10-fold Cross Validation results (mean $\pm$ std) of the accuracy of the GCN  on the graph classification task performed on the ENZYMES/PROTEINS/MNIST/CIFAR10 datasets. We also report the number of parameters relative to the number of parameters in the vanilla model, e.g., $0.37$ means that the number of parameters in \activation model is $37$\% of the vanilla one. We set the best results to bold and underline the second best result. In addition, if the result of \activation model is better than the SGC model, we put $*$ next to the result. %Due to the large computational cost implied by the size of the image datasets we were unable to produce confidence intervals for the provided results.
} \label{tab:res-tu}
\def\arraystretch{1.0}
% \scriptsize
\begin{center}\resizebox{\columnwidth}{!}{
\begin{tabular}{ccccccccc}
\hline
\hline
& \multicolumn{2}{c}{ENZYMES} &  \multicolumn{2}{c}{Proteins} & \multicolumn{2}{c}{MNIST} & \multicolumn{2}{c}{CIFAR10} \\
& ACC. & Params. & ACC. & Params. & ACC. & Params. & ACC. & Params.\\
\hline
Vanilla & $\underline{66.50 \pm 8.71}$ & $1.00$ &  $\mathbf{76.73 \pm 3.85}$ & $1.00$ &$\mathbf{90.77}$ & $1.00$ & $\mathbf{52.04}$ & $1.00$  \\ 

SGC & $63.67 \pm 8.06$ & $0.37$ & $67.65 \pm 2.21$ & $0.38$ & $24.48$ & $0.36$ & $27.90$ & $0.36$\\ \hline

\expander -$50$\% & $64.83 \pm 8.64$ & $0.57$ &  $76.36 \pm 3.43$ & $0.57$ & $\underline{90.75}$ & $0.57$ & $\underline{50.69}$ & $0.57$\\

\expander -$10$\% & $66.33 \pm 6.78$ & $0.22$ &  $\underline{76.55 \pm 1.90}$ & $0.22$ &  $89.00$ & $0.23$ & $50.27$ & $0.23$ \\

\activation & $\mathbf{66.67 \pm 6.71}*$ & $0.37$  & $75.92 \pm 2.88*$ & $0.38$ &$83.84*$ & $0.36$ & $48.31*$ & $0.36$\\
\hline
\hline
\end{tabular}}\\
\end{center}
\end{table*}

\paragraph{Graph Classification.} Table \ref{tab:res-tu} shows the experiment results of the vanilla GCN model and its \expander~and \activation %, \activation and \simple 
variants on the ENZYMES, PROTEINS, MNIST and CIFAR datasets for graph classification. One direct observation  is that the \expander GCN models, even at $10\%$ density, perform on par with the vanilla models. Surprisingly, the same is true for the \activation model on the ENZYMES and PROTEINS datasets.
The SGC performs, often significantly, worse than the \activation model, especially on the computer vision MNIST and CIFAR datasets.
Additional experiments on the GIN architecture can be found in Appendix  Table \ref{tab:res-gin-gclass}. They reaffirm the conclusions drawn on the GCN model on graph classification in this section. 

\begin{table*}[t]
\caption{Results of the GCN/GIN/GraphSage/PNA models on graph regression for the ZINC dataset. The format follows Table~\ref{tab:res-tu}.} \label{tab:other-model}
\def\arraystretch{1.0}
% \scriptsize
\begin{center}\resizebox{\columnwidth}{!}{
\begin{tabular}{ccccccccc}
\hline
\hline
& \multicolumn{2}{c}{GCN} & \multicolumn{2}{c}{GIN} & \multicolumn{2}{c}{GraphSage} & \multicolumn{2}{c}{PNA}\\
& MAE & Params. &  MAE  & Params. &  MAE & Params.  &  MAE & Params. \\
\hline
Vanilla & $\mathbf{0.3823}$ & $1.00$ & $\underline{0.4939}$ & $1.00$ & $\mathbf{0.4530}$ & $1.00$ & $\mathbf{0.3180}$ & $1.00$ \\ 

SGC & $0.6963$ & $0.35$ & --- & --- & --- & --- & --- & --- \\ \hline

\expander -$50$\% & $\underline{0.3856}$ & $0.57$ & $0.5274$ & $0.51$ & $\underline{0.4580}$ & $0.54$ & $\underline{0.3380}$ & $0.51$ \\

\expander -$10$\% & $0.3958$ & $0.22$ & $\mathbf{0.4888}$ & $0.12$ & $0.4720$ & $0.17$ & $0.3800$ & $0.12$ \\

\activation & $0.5855*$ & $0.13$ & $0.5220$ & $0.01$ & $0.4910$ & $0.07$ & $0.4490$ & $0.02$ \\
\hline
\hline
\end{tabular}}\\
\end{center}
\end{table*}

\paragraph{Graph Regression.} In Table \ref{tab:other-model} the Mean Absolute Error (MAE) of our studied and proposed models on the ZINC dataset for graph regression is displayed. 
Similar to the graph classification task, the \expander GCN and \expander GraphSage models match the performance of their corresponding vanilla models, regardless of their densities. 
The performance of the \expander GIN and \expander PNA models exhibits greater variance across the different densities, especially in the case of the PNA models the performance increases as the network gets denser indicating that the density of the {\itshape Update} step does positively contribute to the model performance of the PNA for the task of graph regression on the ZINC dataset. The \activation models perform worse than their \expander~counterparts on this task, confirming the insight from the results of the \expander GNNs that the linear transform in the {\itshape Update} step does improve performance in this graph regression task. Again we see that \activation GCNs outperform the SGC benchmark in this set of experiments. 
    
Hence, for the task of graph regression we observe that both the linear transformation and non-linear activation function in the {\itshape Update} step have a small, positive impact on model performance.

\begin{table*}[t]
\caption{10-fold Cross Validation results (mean $\pm$ std) of the accuracy of the GCN  on node classification for the  CORA/CiteSeer/PubMed/OGBN-Arxiv datasets. The format follows Table~\ref{tab:res-tu} in the main text.}
\label{tab:res-nclass}
\def\arraystretch{1.0}
% \scriptsize
\begin{center}
\resizebox{\columnwidth}{!}{
\begin{tabular}{ccccccccc}
\hline
\hline
& \multicolumn{2}{c}{Cora} & \multicolumn{2}{c}{CiteSeer} & \multicolumn{2}{c}{PubMed} & \multicolumn{2}{c}{OGBN-Arxiv}\\
& ACC. & Params. & ACC. & Params. & ACC. & Params. & ACC. & Params.\\
\hline
Vanilla & $80.54 \pm 0.44$ & $1.00$ & $\underline{69.50 \pm 0.19}$ & $1.00$ & $\underline{79.04 \pm 0.12}$ & $1.00$ & $\underline{71.22 \pm 0.76}$ & $1.00$ \\ 

SGC & $80.40 \pm 0.00$ & $0.03$ & $\mathbf{72.70 \pm 0.00}$ & $0.02$ & $78.90 \pm 0.00$ & $0.01$ & $66.53 \pm 0.07$ & $0.05$ \\ \hline

\expander -$50$\% & $\underline{80.42 \pm 0.28}$ & $0.50$ & $69.43 \pm 0.34$ & $0.50$ & $\mathbf{79.34 \pm 0.28}$ & $0.50$ & $\mathbf{71.42 \pm 0.55}$ & $0.56$ \\

\expander -$10$\% & $\mathbf{80.59 \pm 0.64}$ & $0.10$ & $68.68 \pm 0.73$ & $0.10$ & $78.95 \pm 0.63$ & $0.11$ & $70.70 \pm 0.42$ & $0.19$ \\

\activation & $80.40 \pm 0.00$ & $0.03$ & $\mathbf{72.70 \pm 0.00}$ & $0.02$ & $78.90 \pm 0.00$ & $0.01$ & $68.29 \pm 0.13$ & $0.05$ \\
\hline
\hline
\end{tabular}
}\\
\end{center}
\end{table*}

\paragraph{Node Classification} Results from the node classification experiments on four citation graphs (CORA, CITESEER, PUBMED and ogbn-arxiv) can be found in Table \ref{tab:res-nclass}. For medium-sized datasets such as CORA, CITESEER and PUBMED, we have the same observation as for the graph classification and graph regression tasks, the \expander~models, regardless of their sparsity, are performing on par with the vanilla ones. 
The \activation models also perform as well as or even better than (on CITESEER) the vanilla model. 
The performance of the GCN  \activation model and SGC is equally good across all three datasets.
These conclusions remain true for the large-scale dataset ogbn-arxiv with 169,343 nodes and 1,166,243 edges. The {\itshape ExpanderGCNs} are on par with the vanilla GCN while the \activation model and SGC perform slightly worse. However, the training time of \activation model and SGC is five times faster than that of the \expander ~ and vanilla models. %We are also able to observe the influence of the choice of activation function. 
The \activation model outperforms the SGC. Additional experiments on the GIN architecture can be found in Appendix  Table \ref{tab:gin-nclass}. %, while the other two \activation models do worse. 

We observe that in the node classification task both the linear transformation and the non-linear activation function offer no benefit for the medium scale datasets. For the large-scale dataset we find that the linear transformation can be sparsified heavily without a loss in performance, but deleting it entirely does worsen model performance.

%%%%%%%%%%%%%%%%%%%%%%%%%%%%%%%%%%%%%%%%%%%%
%%%%%%############################%%%%%%%%%%
%%%%%%%%%%%%%%%%%%%%%%%%%%%%%%%%%%%%%%%%%%%%
\section{Conclusion }\label{sec:conclusion}
With extensive experiments across different GNN models and graph learning tasks, we are able to confirm that the {\itshape Update} step can be sparsified heavily without a significant performance cost. 
In fact for six of the nine tested datasets across a variety of tasks we found that the linear transform can be removed entirely without a loss in performance, i.e., the \activation models performed on par with their vanilla counterparts. The \activation GCN model consistently outperformed the SGC model and especially in the computer vision datasets we witnessed that the activation functions seem to be crucial for good model performance accounting for an accuracy difference of up to 59\%.
These findings partially support the hypothesis by \citet{wu2019simplifying} that the {\itshape Update} step can be simplified significantly without a loss in performance. Contrary to  \citet{wu2019simplifying} we find that the nonlinear activation functions result in a significant accuracy boost and the linear transformation in the {\itshape Update} step can be removed or heavily sparsified. 

The \activation GNN is an effective and simple benchmark model framework for any message passing neural network. It enables practitioners to test whether they can cut the large amount of model parameters used in the linear transform of the {\itshape Update} steps. If the linear transform does contribute positively to the model's performance then the \expander GNNs provide a model class of tuneable sparsity which allows efficient parameter usage.

\section*{Acknowledgements}
The work of Dr. Johannes Lutzeyer and Prof. Michalis Vazirgiannis is supported by the ANR chair AML-HELAS (ANR-CHIA-0020-01).

\newpage
\bibliography{reference}

\begin{thebibliography}{39}
\providecommand{\natexlab}[1]{#1}
\providecommand{\url}[1]{\texttt{#1}}
\expandafter\ifx\csname urlstyle\endcsname\relax
  \providecommand{\doi}[1]{doi: #1}\else
  \providecommand{\doi}{doi: \begingroup \urlstyle{rm}\Url}\fi

\bibitem[Blalock et~al.(2020)Blalock, Ortiz, Frankle, and Guttag]{Blalock2020}
Davis Blalock, Jose Javier~Gonzalez Ortiz, Jonathan Frankle, and John Guttag.
\newblock What is the state of neural network pruning?
\newblock In \emph{Conference on Machine Learning and Systems}, 2020.

\bibitem[B\"olcskei et~al.(2019)B\"olcskei, Grohs, Kutyniok, and
  Petersen]{Bolcskei2019}
Helmut B\"olcskei, Philipp Grohs, Gitta Kutyniok, and Philipp Petersen.
\newblock Optimal approximation with sparsely connected deep neural networks.
\newblock \emph{SIAM Journal on Mathematics of Data Science}, pp.\  8--45,
  2019.

\bibitem[Bourely et~al.(2017)Bourely, Boueri, and Choromonski]{Bourely2017}
Alfred Bourely, John~Patrick Boueri, and Krzysztof Choromonski.
\newblock Sparse neural networks topologies.
\newblock \emph{arXiv:1706.05683}, 2017.

\bibitem[CEREBRAS~SYSTEMS(2021)]{Cerebras2021}
INC CEREBRAS~SYSTEMS.
\newblock Cerebras white paper 3: Cerebras systems: Achieving industry best ai
  performance through a systems approach.
\newblock
  \url{https://cerebras.net/wp-content/uploads/2021/04/Cerebras-CS-2-Whitepaper.pdf},
  2021.
\newblock accessed May 2021.

\bibitem[Corso et~al.(2020)Corso, Cavalleri, Beaini, Li{\`o}, and
  Veli{\v{c}}kovi{\'c}]{corso2020principal}
Gabriele Corso, Luca Cavalleri, Dominique Beaini, Pietro Li{\`o}, and Petar
  Veli{\v{c}}kovi{\'c}.
\newblock Principal neighbourhood aggregation for graph nets.
\newblock \emph{arXiv:2004.05718}, 2020.

\bibitem[Duvenaud et~al.(2015)Duvenaud, Maclaurin, Iparraguirre, Bombarell,
  Hirzel, Aspuru-Guzik, and Adams]{duvenaud2015convolutional}
David~K Duvenaud, Dougal Maclaurin, Jorge Iparraguirre, Rafael Bombarell,
  Timothy Hirzel, Al{\'a}n Aspuru-Guzik, and Ryan~P Adams.
\newblock Convolutional networks on graphs for learning molecular fingerprints.
\newblock In \emph{Advances in neural information processing systems}, pp.\
  2224 -- 2232, 2015.

\bibitem[Dwivedi et~al.(2020{\natexlab{a}})Dwivedi, Joshi, Laurent, Bengio, and
  Bresson]{dwivedi2020benchmarking}
Vijay~Prakash Dwivedi, Chaitanya~K. Joshi, Thomas Laurent, Yoshua Bengio, and
  Xavier Bresson.
\newblock Benchmarking graph neural networks.
\newblock \emph{arXiv:2003.00982}, 2020{\natexlab{a}}.

\bibitem[Dwivedi et~al.(2020{\natexlab{b}})Dwivedi, Joshi, Laurent, Bengio, and
  Bresson]{dwivedi2020code}
Vijay~Prakash Dwivedi, Chaitanya~K Joshi, Thomas Laurent, Yoshua Bengio, and
  Xavier Bresson.
\newblock Benchmarking graph neural networks.
\newblock \url{https://github.com/graphdeeplearning/benchmarking-gnns},
  2020{\natexlab{b}}.
\newblock accessed May 2020.

\bibitem[Frankle \& Carbin(2019)Frankle and Carbin]{Frankle2019}
Jonathan Frankle and Michael Carbin.
\newblock The lottery ticket hypothesis: Finding sparse, trainable neural
  networks.
\newblock In \emph{7th International Conference on Learning Representations
  (ICLR)}, 2019.

\bibitem[Frankle et~al.(2021)Frankle, Dziugaite, Roy, and Carbin]{Frankle2021}
Jonathan Frankle, Gintare~Karolina Dziugaite, Daniel~M Roy, and Michael Carbin.
\newblock Pruning neural networks at initialization: Why are we missing the
  mark?
\newblock In \emph{9th International Conference on Learning Representations
  (ICLR)}, 2021.

\bibitem[Gama et~al.(2020)Gama, Ribeiro, and Bruna]{fern2019stability}
Fernando Gama, Alejandro Ribeiro, and Joan Bruna.
\newblock Stability of graph neural networks to relative perturbations.
\newblock In \emph{International Conference on Acoustics, Speech, and Signal
  Processing (ICASSP)}, pp.\  9070 -- 9074. IEEE, 2020.

\bibitem[Gilmer et~al.(2017)Gilmer, Schoenholz, Riley, Vinyals, and
  Dahl]{gilmer2017neural}
Justin Gilmer, Samuel~S Schoenholz, Patrick~F Riley, Oriol Vinyals, and
  George~E Dahl.
\newblock Neural message passing for quantum chemistry.
\newblock In \emph{Proceedings of the 34th International Conference on Machine
  Learning (ICML)}, pp.\  1263 -- 1272, 2017.

\bibitem[Griffa et~al.(2017)Griffa, Ricaud, Benzi, Bresson, Daducci,
  Vandergheynst, Thiran, and Hagmann]{griffa2017transient}
Alessandra Griffa, Benjamin Ricaud, Kirell Benzi, Xavier Bresson, Alessandro
  Daducci, Pierre Vandergheynst, Jean-Philippe Thiran, and Patric Hagmann.
\newblock Transient networks of spatio-temporal connectivity map communication
  pathways in brain functional systems.
\newblock \emph{NeuroImage}, pp.\  490 -- 502, 2017.

\bibitem[Hamilton et~al.(2017)Hamilton, Ying, and
  Leskovec]{hamilton2017inductive}
William~L. Hamilton, Rex Ying, and Jure Leskovec.
\newblock Inductive representation learning on large graphs.
\newblock In \emph{Proceedings of the 31st International Conference on Neural
  Information Processing Systems (NIPS)}, pp.\  1025 -- 1035. Curran Associates
  Inc., 2017.

\bibitem[Han et~al.(2015)Han, Pool, Tran, and Dally]{Han2015}
Song Han, Jeff Pool, John Tran, and William Dally.
\newblock Learning both weights and connections for efficient neural network.
\newblock In \emph{Advances in neural information processing systems}, pp.\
  1135--1143, 2015.

\bibitem[Hoefler et~al.(2021)Hoefler, Alistarh, Ben-Nun, Dryden, and
  Peste]{Hoefler2021}
Torsten Hoefler, Dan Alistarh, Tal Ben-Nun, Nikoli Dryden, and Alexandra Peste.
\newblock Sparsity in deep learning: Pruning and growth for efficient inference
  and training in neural networks.
\newblock \emph{arXiv preprint arXiv:2102.00554}, 2021.

\bibitem[Hoory et~al.(2006)Hoory, Linial, and Wigderson]{Hoory2006ExpanderGA}
Shlomo Hoory, Nathan Linial, and Avi Wigderson.
\newblock Expander graphs and their applications.
\newblock \emph{Bulletin of the American Mathematical Society}, pp.\  439 --
  561, 2006.

\bibitem[Hu et~al.(2020)Hu, Fey, Zitnik, Dong, Ren, Liu, Catasta, and
  Leskovec]{hu2020open}
Weihua Hu, Matthias Fey, Marinka Zitnik, Yuxiao Dong, Hongyu Ren, Bowen Liu,
  Michele Catasta, and Jure Leskovec.
\newblock Open graph benchmark: Datasets for machine learning on graphs.
\newblock \emph{arXiv preprint arXiv:2005.00687}, 2020.

\bibitem[Irwin et~al.(2012)Irwin, Sterling, Mysinger, Bolstad, and
  Coleman]{irwin2012zinc}
John~J Irwin, Teague Sterling, Michael~M Mysinger, Erin~S Bolstad, and Ryan~G
  Coleman.
\newblock Zinc: a free tool to discover chemistry for biology.
\newblock \emph{Journal of chemical information and modeling}, pp.\  1757 --
  1768, 2012.

\bibitem[Kepner \& Robinett(2019)Kepner and Robinett]{Robinett2019}
Jeremy Kepner and Ryan Robinett.
\newblock Radix-net: Structured sparse matrices for deep neural networks.
\newblock In \emph{2019 IEEE International Parallel and Distributed Processing
  Symposium Workshops (IPDPSW)}, pp.\  268 -- 274. IEEE, 2019.

\bibitem[Kersting et~al.(2016)Kersting, Kriege, Morris, Mutzel, and
  Neumann]{KKMMN2016}
Kristian Kersting, Nils~M. Kriege, Christopher Morris, Petra Mutzel, and Marion
  Neumann.
\newblock Benchmark data sets for graph kernels, 2016.
\newblock \url{http://graphkernels.cs.tu-dortmund.de}.

\bibitem[Kipf \& Welling(2017)Kipf and Welling]{kipf2016semi}
Thomas~N. Kipf and Max Welling.
\newblock Semi-supervised classification with graph convolutional networks.
\newblock In \emph{5th International Conference on Learning Representations
  (ICLR)}, 2017.

\bibitem[Knyazev et~al.(2019)Knyazev, Taylor, and
  Amer]{knyazev2019understanding}
Boris Knyazev, Graham~W Taylor, and Mohamed Amer.
\newblock Understanding attention and generalization in graph neural networks.
\newblock In \emph{Advances in Neural Information Processing Systems 32}, pp.\
  4202 -- 4212. Curran Associates, Inc., 2019.

\bibitem[Lee et~al.(2019)Lee, Ajanthan, and Torr]{Lee2019}
Namhoon Lee, Thalaiyasingam Ajanthan, and Philip~HS Torr.
\newblock {SNIP}: Single-shot network pruning based on connection sensitivity.
\newblock In \emph{7th International Conference on Learning Representations
  (ICLR)}, 2019.

\bibitem[Lubotzky(2012)]{lubotzky2011expander}
Alexander Lubotzky.
\newblock Expander graphs in pure and applied mathematics.
\newblock \emph{Bulletin of the American Mathematical Society}, pp.\  113 --
  162, 2012.

\bibitem[McDonald \& Shokoufandeh(2019)McDonald and Shokoufandeh]{McDonald2019}
Andrew~WE McDonald and Ali Shokoufandeh.
\newblock Sparse super-regular networks.
\newblock In \emph{18th IEEE International Conference On Machine Learning And
  Applications (ICMLA)}, pp.\  1764 -- 1770. IEEE, 2019.

\bibitem[Monti et~al.(2019)Monti, Frasca, Eynard, Mannion, and
  Bronstein]{monti2019fake}
Federico Monti, Fabrizio Frasca, Davide Eynard, Damon Mannion, and Michael~M
  Bronstein.
\newblock Fake news detection on social media using geometric deep learning.
\newblock \emph{arXiv:1902.06673}, 2019.

\bibitem[NVIDIA(2020)]{Nvidia2020}
NVIDIA.
\newblock Nvidia white paper: Nvidia a100 tensor core gpu architecture.
\newblock
  \url{https://images.nvidia.com/aem-dam/en-zz/Solutions/data-center/nvidia-ampere-architecture-whitepaper.pdf},
  2020.
\newblock accessed May 2021.

\bibitem[Prabhu et~al.(2018)Prabhu, Varma, and Namboodiri]{prabhu2018deep}
Ameya Prabhu, Girish Varma, and Anoop Namboodiri.
\newblock Deep expander networks: Efficient deep networks from graph theory.
\newblock In \emph{Proceedings of the European Conference on Computer Vision
  (ECCV)}, pp.\  20 -- 35, 2018.

\bibitem[Salha et~al.(2019)Salha, Hennequin, and Vazirgiannis]{Salha2019}
Guillaume Salha, Romain Hennequin, and Michalis Vazirgiannis.
\newblock Keep it simple: Graph autoencoders without graph convolutional
  networks.
\newblock In \emph{Workshop on Graph Representation Learning at the 33rd
  Conference on Neural Information Processing Systems (NeurIPS)}, 2019.

\bibitem[Sen et~al.(2008)Sen, Namata, Bilgic, Getoor, Galligher, and
  Eliassi-Rad]{sen2008collective}
Prithviraj Sen, Galileo Namata, Mustafa Bilgic, Lise Getoor, Brian Galligher,
  and Tina Eliassi-Rad.
\newblock Collective classification in network data.
\newblock \emph{AI magazine}, pp.\  93 -- 93, 2008.

\bibitem[Strategy(2020)]{Graphcore2020}
Moor Insights~\& Strategy.
\newblock Graphcore white paper: The graphcore second generation ipu.
\newblock
  \url{https://www.graphcore.ai/hubfs/MK2-%20The%20Graphcore%202nd%20Generation%20IPU%20Final%20v7.14.2020.pdf?hsLang=en},
  2020.
\newblock accessed May 2021.

\bibitem[Tanaka et~al.(2020)Tanaka, Kunin, Yamins, and
  Ganguli]{tanaka2020pruning}
Hidenori Tanaka, Daniel Kunin, Daniel~LK Yamins, and Surya Ganguli.
\newblock Pruning neural networks without any data by iteratively conserving
  synaptic flow.
\newblock \emph{arXiv:2006.05467}, 2020.

\bibitem[Wang et~al.(2020{\natexlab{a}})Wang, Zhang, and Grosse]{Wang2020}
Chaoqi Wang, Guodong Zhang, and Roger Grosse.
\newblock Picking winning tickets before training by preserving gradient flow.
\newblock In \emph{8th International Conference on Learning Representations
  (ICLR)}, 2020{\natexlab{a}}.

\bibitem[Wang et~al.(2020{\natexlab{b}})Wang, Shen, Huang, Wu, Dong, and
  Kanakia]{wang2020mcirosoft}
Kuansan Wang, Zhihong Shen, Chiyuan Huang, Chieh-Han Wu, Yuxiao Dong, and
  Anshul Kanakia.
\newblock Microsoft academic graph: When experts are not enough.
\newblock \emph{Quantitative Science Studies}, 1\penalty0 (1):\penalty0
  396--413, 2020{\natexlab{b}}.

\bibitem[Wu et~al.(2019)Wu, Souza~Jr, Zhang, Fifty, Yu, and
  Weinberger]{wu2019simplifying}
Felix Wu, Amauri~H Souza~Jr, Tianyi Zhang, Christopher Fifty, Tao Yu, and
  Kilian~Q Weinberger.
\newblock Simplifying graph convolutional networks.
\newblock In \emph{Proceedings of the 36th International Conference on Machine
  Learning (ICML)}, 2019.

\bibitem[Wu et~al.(2020)Wu, Pan, Chen, Long, Zhang, and
  Philip]{wu2020comprehensive}
Zonghan Wu, Shirui Pan, Fengwen Chen, Guodong Long, Chengqi Zhang, and S~Yu
  Philip.
\newblock A comprehensive survey on graph neural networks.
\newblock \emph{IEEE Transactions on Neural Networks and Learning Systems},
  pp.\  1 -- 21, 2020.

\bibitem[Xu et~al.(2019)Xu, Hu, Leskovec, and Jegelka]{xu2018powerful}
Keyulu Xu, Weihua Hu, Jure Leskovec, and Stefanie Jegelka.
\newblock How powerful are graph neural networks?
\newblock In \emph{7th International Conference on Learning Representations
  (ICLR)}, 2019.

\bibitem[Yao et~al.(2019)Yao, Mao, and Luo]{yao2019graph}
Liang Yao, Chengsheng Mao, and Yuan Luo.
\newblock Graph convolutional networks for text classification.
\newblock In \emph{Proceedings of the AAAI Conference on Artificial
  Intelligence}, pp.\  7370 -- 7377, 2019.

\end{thebibliography}
\bibliographystyle{iclr2022_workshop}
\newpage
\appendix
\section*{Appendix}

\section{Motivation and Implementation of the ExpanderGNNs} \label{app:mot_imp_exgnn}

\paragraph{Motivation of the Expander Linear Layer}
From our random expander sampling scheme discussed in Section \ref{sec:model} we sample bipartite graphs with good expansion properties, which are commonly discussed in the field of error correcting codes under the name ``lossless expanders''  \citep[pp.~517-522]{Hoory2006ExpanderGA}. Expander graphs can be informally defined to be highly connected and sparse graphs \citep{lubotzky2011expander}. 
They are successfully applied in communication networks where communication comes at a certain cost and is to be used such that messages are spread across the network efficiently \citep{lubotzky2011expander}. 
Equally, in a neural network each parameter (corresponding to an edge in the neural network architecture) incurs a computational cost and is placed to optimise the overall performance of the neural network architecture. Therefore, the use of expander graphs in the design of neural network architectures is conceptually well-motivated.

% Designing neural network architectures using the concept of expander graphs leads to neural network architectures, where fewer edges are placed in such a way that the overall computational structure remains highly connected.
In \citet{Bolcskei2019}  the connectedness of a sparse neural network architecture was linked to the complexity of a given function class which can be approximated by sparse neural networks. Hence, utilising neural network parameters to optimise the connectedness of the network maximises the expressivity of the neural network. 
In \citet{Robinett2019} and \citet{Bourely2017} the connectedness of the neural network architecture graph was linked -- via the path-connectedness and the graph Laplacian eigenvalues -- to the performance of neural network architectures. Therefore, for both the expressivity of the neural network and its performance, the connectedness, which is optimised in expander graphs, is a parameter of interest. %Expander graphs utilise a given budget of edges, corresponding to parameters in the neural network, to form graphs which are highly connected. 

\paragraph{Implementation of Expander Linear Layer} \label{sec:expander_implementation}
The most straightforward way of implementing the expander linear layer is to store the weight matrix $\bm{W}$ as a sparse matrix. 
Sparse matrix multiplications can be accelerated on several processing units released in 2020 and 2021 such as the 
Sparse Linear Algebra Compute (SLAC) cores used in the Cerebras WSE-2 \citep{Cerebras2021}, the
  Intelligence Processing Unit (IPU) produced by Graphcore \citep{Graphcore2020}
and the NVIDIA A100 Tensor Core GPU \citep{Nvidia2020}.
%However, due to the known issue of inefficiency of hardware acceleration on sparse matrices \cite{wen2016learning}, 
However, since we ran experiments on a NVIDIA RTX 2060 GPU, %\todo{XXX} GPU, 
we use masks in our implementation, similar to those of several existing pruning algorithms, to achieve the sparsification. A mask $\bm{M}\in\{0,1\}^{|\mathbb{S}_1|\times |\mathbb{S}_2|}$ is of the same dimension as weight matrix and $M_{u,v} = 1$ if and only if $(u,v) \in \mathbb{E}'.$ An entrywise multiplication, denoted by $\odot,$ is then applied to the mask and the weight matrix so that undesired parameters in the weight matrix are removed. To illustrate this alteration we use the following matrix representation of the GCN's model equation, 
\begin{equation}\label{eq:weight}
    \bm{H}^{(L)} = \sigma\left(\hat{\bm{A}}\ldots\sigma\left(\hat{\bm{A}}\bm{H}^{(1)}\bm{W}^{(1)}\right)\ldots\bm{W}^{(L)}\right).
\end{equation}
In our implementation the \expander GCN has the following model equation,
\begin{equation}\label{eq:expander}
    \bm{H}^{(L)} = \sigma\left(\hat{\bm{A}}\ldots\sigma\left(\hat{\bm{A}}\bm{H}^{(1)}\bm{M}^{(1)}\odot\bm{W}^{(1)}\right)\ldots\bm{M}^{(L)}\odot\bm{W}^{(L)}\right).
\end{equation}

\section{General Settings and Baselines} \label{sec:exp_general_setting}

\paragraph{Considered GNNs} Throughout this section we refer to the standard, already published, architectures as ``vanilla'' architectures. %For each dataset on each task, we
We compare the performance of %of evaluate with same settings 
the vanilla GNN models, the \expander GNN models with different densities ($10\%$ and $50\%$), the \activation GNN model (we report the best result obtained from the ReLU, PReLU and Tanh activation functions), % with different activation functions (ReLU, PReLU, Tanh), 
as well as the SGC for the GCN models. 
To ensure that our inference is not specific to a certain GNN architecture only, we evaluate the performance across 4 representative GNN models of the literature state-of-the-art. The considered models are the Graph Convolutional Network (GCN, \citet{kipf2016semi}), the Graph Isomorphism Network (GIN, \citet{xu2018powerful}), the GraphSage Network \citet{hamilton2017inductive}, and the Principle Neighborhood Aggregation (PNA, \citet{corso2020principal}).  
The precise model equations of our proposed architectures applied to these GNNs can be found in Table \ref{tab:model_implementations}.
We make the following additional remarks with regard to this table:
% \noindent We make the following remarks with regard to Table \ref{tab:model_implementations}:
   \begin{enumerate}
       \item In the model equations of the GCN, $d_i$ denotes the degree of node $i.$
       \item In the model equations of the GIN, $\epsilon$ is a learnable ratio added explicitly to the central node's own representation.
       \item In the model equations of the PNA, $\bigoplus$ corresponds to an operator formed by taking the tensor product of a vector containing three scaler functions and four aggregator functions, resulting in a tensor indexed by $i,s,a,$ where the index $i$ corresponds to the currently considered node, $s$ corresponds to the scaler dimension and $a$ indexes the aggregator dimension. For more details see \citet{corso2020principal}.

   \end{enumerate}

\begin{table*}[t!]
\centering
\def\arraystretch{1.2}
\footnotesize
\caption{Model Equations of the Vanilla, \expander ~and \activation GNNs.}
\label{tab:model_implementations}
\resizebox{\columnwidth}{!}{
\begin{tabular}{lllll}
\hline
\hline
\multicolumn{2}{l}{Model} &  Aggregation  &  Update  \\
\hline
\multirow{2}{*}{GCN} & Vanilla/\expander & \multirow{2}{*}{$\bm{m}_i^{(l)} = \frac{1}{\sqrt{d_i}} \sum_{j\in\mathcal{N}(i)} \bm{h}_j^{(l)} \frac{1}{\sqrt{d_j}}$} &  $\bm{h}_i^{(l+1)} = \sigma(\bm{m}_i^{(l)}\bm{M}^{(l)}\odot\bm{W}^{(l)})$\\
% \cline{2-2} \cline{4-4}
 & \activation  &  &  $\bm{h}_i^{(l+1)} = \sigma\left(\bm{m}_i^{(l)}\right)$ \\ 
\cline{1-4}
\multirow{2}{*}{GIN} & Vanilla/\expander & \multirow{2}{*}{$\bm{m}_i^{(l)} = (1+\epsilon)\bm{h}_i^{(l)} + \sum_{j\in\mathcal{N}(i)} \bm{h}_j^{(l)}$} & $\bm{h}_i^{(l+1)} = \sigma(\bm{m}_i^{(l)}\bm{M}^{(l)}\odot\bm{W}^{(l)})$ \\
% \cline{2-2} \cline{4-4}
 & \activation &  &  $\bm{h}_i^{(l+1)} = \sigma\left(\bm{m}_i^{(l)}\right)$ \\ 
\cline{1-4}
\multirow{2}{*}{GraphSage} & Vanilla/\expander & $ \bm{m}_i^{(l)} = \mathrm{CONCAT}\left( \bm{h}_i^{(l)},  \mathrm{MAX}_{j\in\mathcal{N}(i)}\sigma(\bm{h}_j^{(l)}\bm{M}_1^{(l)}\odot\bm{W}_1^{(l)})\right)$ 
& $\bm{h}_i^{(l+1)} = \frac{\sigma(\bm{m}_i^{(l)}\bm{M}_2^{(l)}\odot\bm{W}_2^{(l)})}{\|\sigma(\bm{m}_i^{(l)}\bm{M}_2^{(l)}\odot\bm{W}_2^{(l)})\|_2}$ \\
% \cline{2-4}
 & \activation & $\bm{m}_i^{(l)} = \bm{h}_i^{(l)} + \mathrm{MAX}_{j\in\mathcal{N}(i)}\sigma(\bm{h}_j^{(l)})$ 
 & $
\bm{h}_i^{(l+1)} = \frac{\sigma(\bm{m}_i^{(l)})}{\|\sigma(\bm{m}_i^{(l)})\|_2}$ \\ 
\cline{1-4}
\multirow{2}{*}{PNA} & Vanilla/\expander & $\bm{m}_i^{(l)} = \mathrm{CONCAT}_{s,a}\left(\bigoplus_{j\in\mathcal{N}(i)}\bm{h}_j^{(l)}\right)$ &
$\bm{h}_i^{(l+1)} = \sigma(\bm{m}_i^{(l)}\bm{M}^{(l)}\odot\bm{W}^{(l)})$ \\
% \cline{2-4}
& \activation &  $ \bm{m}_i^{(l)} = \frac{1}{12} \sum_{s,a}\left[\bigoplus_{j\in\mathcal{N}(i)}\bm{h}_j^{(l)}\right]_{i,s,a}$ & $\bm{h}_i^{(l+1)} = \sigma \left( \bm{m}_i^{(l)} \right)$  \\
\hline
\hline
\end{tabular}}
\end{table*}

\begin{table*}[t]
\centering

\caption{Properties of all datasets used in experiments.}
\label{tab:datasets}
% \small
\resizebox{\columnwidth}{!}{
\begin{tabular}{cccccc}%{c|c|c|c|c|c}
\hline
\hline
& Dataset & \#Graphs & \#Nodes  (avg.) & \#Edges  (avg.) & Task \\
\hline
\multirow{ 2}{*}{TU datasets} & ENZYMES & $600$ & $32.63$ & $62.14$ & \multirow{4}{*}{Graph Classification} \\
& PROTEINS & $1113$ & $39.06$ & $72.82$ & \\ 
\cline{1-5}
\multirow{ 2}{*}{Computer Vision} & MNIST & $70000$ & $70.57$ & $282.27$ & \\
& CIFAR$10$ & $60000$ & $117.63$ & $470.53$ & \\ \hline
% Computer & MNIST & $70000$ & $40$-$75$ &  & \\
% Vision& CIFAR$10$ & $150000$ & $85$-$150$ &  & \\ \hline
& ZINC & $12000$ & $23.16$ & $24.92$ & Graph Regression \\ \hline 
\multirow{ 4}{*}{Citations} & CORA & $1$ & $2708$ & $5278$ &\multirow{4}{*}{Node Classification} \\ & CITESEER & $1$ & $3327$ & $4552$ & \\
& PUBMED & $1$ & $19717$ & $44324$ & \\
& ogbn-arxiv & $1$ & $169343$ & $1166243$ \\
\hline
\hline
\end{tabular}
}
\end{table*}

\paragraph{Datasets} We experiment on eleven datasets from areas such as chemistry, social networks, computer vision and academic citation, for three major graph learning tasks. 
For graph classification, we have two TU datasets \citep{KKMMN2016} which are chemical graphs, and two Image datasets (MNIST/CIFAR10) that are constructed from original images following the procedure in \citet{knyazev2019understanding}.
%mentioned in Section~\ref{sec:exp_graph_classification}. 
To perform this conversion they first extract small regions of homogeneous intensity from the images, named ``Superpixels'' \citep{dwivedi2020benchmarking}, and construct a $K$-nearest neighbour graph from these superpixels. % as a graph representation for the corresponding image. 
The technique we implemented to extract superpixels, the choice of $K$ and distance kernel for constructing a nearest neighbour graph are the same as in  \citet{knyazev2019understanding} and \citet{dwivedi2020benchmarking}. 
For graph regression, we consider molecule graphs from the ZINC dataset \citep{irwin2012zinc}. And for node classification, we use four citation datasets \citep{sen2008collective, wang2020mcirosoft, hu2020open}, where the nodes are academic articles linked by citations. 
Details of the used datasets can be found in Table \ref{tab:datasets}, where in the number of nodes and edges column we display average values if the dataset contains multiple graphs.

% In Table~\ref{tab:datasets}, we summarise the statistics of the aforementioned eleven datasets in detail. We display their number of graphs, number of nodes and of edges, as well as the tasks that are performed on them. In the number of nodes and edges column we show the values, or the average of these values if there are multiple graphs.

% % In the \expander GNN models, all MLP layers (except for readout layer in graph level tasks) are replaced by expander linear layers, while they are removed in \activation variants \footnote{Note that sometimes the aggregation function in \activation models is forced to be changed from that of vanilla models or \expander models due to dimension incoherence, i.e. from {\itshape concatenation} in vanilla models to {\itshape sum} or {\itshape mean} in order to maintain feature dimension coherence}. In \simple models, we remove all non-linear activation functions and keep only linear transform layer. If possible, we follow the technique in \cite{wu2019simplifying} to summarise linear transforms in different message-passing layers into one single transformation following the final message-passing step.

\begin{table*}[t]
\caption{10-fold Cross Validation results (mean $\pm$ std) of the accuracy of the GIN on Graph classification for ENZYMES and Proteins, as well as Results for MNIST and CIFAR10. The format follows Table~\ref{tab:res-tu} in the main text.} \label{tab:res-gin-gclass}
\def\arraystretch{1.0}
\scriptsize
\begin{center}
\resizebox{\columnwidth}{!}{
\begin{tabular}{ccccccccccccc}
\hline
\hline
& \multicolumn{2}{c}{ENZYMES}  & \multicolumn{2}{c}{Proteins} &\multicolumn{2}{c}{MNIST} & \multicolumn{2}{c}{CIFAR10}\\
& ACC. & Params.  & ACC. & Params. & ACC. & Params. & ACC. & Params.\\
\hline
Vanilla          & $\mathbf{67.67 \pm 7.68}$ & $1.00$ &  $\mathbf{72.51 \pm 2.39}$ & $1.00$ &  $\underline{90.33}$ & $1.00$ & $\mathbf{42.46}$ & $1.00$\\ 
\hline
\expander -$50$\%  & $\underline{67.00 \pm 6.05}$ & $0.54$ & $70.08 \pm 2.69$ & $0.52$ &  $\mathbf{92.31}$ & $0.56$ & $\underline{40.35}$ & $0.56$\\

\expander -$10$\%  & $65.83 \pm 7.75$ & $0.16$ &  $70.53 \pm 3.96$ & $0.13$ & $88.73$ & $0.20$ & $35.93$ & $0.20$\\

\activation  & $62.83 \pm 7.15$ & $0.10$ & $\underline{72.40 \pm 5.03}$ & $0.08$ &  $79.49$ & $0.11$ & $39.71$ & $0.11$\\
\hline
\hline
\end{tabular}
}\\
\end{center}
\end{table*}

\begin{table*}[t]
\caption{10-fold Cross Validation Results (mean $\pm$ std) of the accuracy of the GIN on node classification for Cora/CiteSeer/PubMed/OGBN-Arxiv. The format follows Table~\ref{tab:res-tu} in the main text.} \label{tab:gin-nclass}
\def\arraystretch{1.0}
% \scriptsize
\begin{center}
\resizebox{\columnwidth}{!}{
\begin{tabular}{ccccccccc}
\hline
\hline
& \multicolumn{2}{c}{Cora} & \multicolumn{2}{c}{CiteSeer} & \multicolumn{2}{c}{PubMed} & \multicolumn{2}{c}{OGBN-Arxiv}\\
& ACC. & Params. & ACC. & Params. & ACC. & Params. & ACC. & Params.\\
\hline
Vanilla & $76.57 \pm 1.36 $ & $1.00$ & $\mathbf{68.33 \pm 0.56}$ & $1.00$ & $76.55 \pm 0.84$ & $1.00$ & $\mathbf{69.37 \pm 0.34}$ & $1.00$ \\ 
\hline

\expander -$50$\% & $77.06 \pm 0.81$ & $0.52$ & $\underline{67.24 \pm 0.49}$ & $0.51$ & $76.06 \pm 1.14$ & $0.51$ & $\underline{69.11 \pm 0.86}$ & $0.59$ \\

\expander -$10$\% & $\underline{77.08 \pm 0.96}$ & $0.13$ & $64.83 \pm 0.49$ & $0.12$ & $\underline{76.91 \pm 0.51}$ & $0.12$ & $68.78 \pm 0.31$ & $0.26$ \\

\activation & $\mathbf{78.65 \pm 0.36}$ & $0.07$ & $66.70 \pm 0.65$ & $0.06$ & $\mathbf{77.46 \pm 0.67}$ & $0.02$ & $64.10 \pm 0.32$ & $0.08$ \\
\hline
\hline
\end{tabular}
}\\
\end{center}
\end{table*}

\paragraph{Experimentation Details} %Settings
Since we aim to observe the performance of our benchmark models independent of the GNN choice we use the model hyperparameters found to yield a fair comparison of GNN models in \citet{dwivedi2020benchmarking}.
% Other experiment details, such as the choice of loss functions for different tasks, dataset splits as well as the extact message-passing formulation of the models we studied and their variants can be found in Appendix~\ref{appx:exp-setting}.  
% In order to ensure a fair comparison across different GNN models, we follow the recent benchmark proposed in \cite{dwivedi2020benchmarking}. 
Specifically, %we use their datasets on computer vision (MNIST/CIFAR10) and chemistry(TU datasets/ZINC dataset); 
we follow the same training procedure, such as train/valid/test dataset splits, choice of optimiser, learning rate decay scheme, as well as the same hyperparameters, such as initial learning rate, hidden feature dimensions and number of GNN layers. We also implement the same normalisation tricks such as adding batch normalisation after non-linearity of each {\itshape Update} step. Their setting files (training procedure/hyperparameters) are made public and can be found in \citet{dwivedi2020code}.  %\href{https://github.com/graphdeeplearning/benchmarking-gnns/tree/master/configs}{repository}. 
For the node classification task on citation datasets, we follow the settings from \citet{wu2019simplifying}. Our experiments found that the node classification task on citation graphs of small to medium size can be easily overfit and model performances heavily depend on the choice of hyperparameters. Using the same parameters with \citet{wu2019simplifying}, such as learning rate, number of training epochs and number of GNN layers, helps us achieve similar results with the paper on the same model, which allows a fair comparison between the proposed \activation models and the SGC.

\paragraph{Additional Results on the GIN architecture} In Tables \ref{tab:res-gin-gclass} and \ref{tab:gin-nclass} we display additional experiments on the GIN architecture on graph classification and node classification, respectively. These results further support the conclusions drawn in the main paper. 

%%%%%%%%%%%%%%%%%%%%%%%%%%%%%%%%%%%%%%%%
% %%MOVED TO APPENDIX FOR AAAI:
%%%%%%%%%%%%%%%%%%%%%%%%%%%%%%%%%%%%%%%%
% \subsection{Loss functions for different tasks}
\paragraph{Loss functions} After $L$ message-passing iterations, we obtain 
 \begin{displaymath}
\bm{H}^{(L)}=\left[\bm{h}_1^{(L)}, \ldots, \bm{h}_n^{(L)}\right]^{\mathrm{T}} \in\mathbb{R}^{n\times p},
 \end{displaymath}
as the final node embedding, where we denote $p$ as its feature dimension. Depending on the downstream task, we either keep working with $\bm{H}^{(L)}$ or  
construct a graph-level representation $\bm{g}$ from $\bm{H}^{(L)},$
$$ \bm{g} = \frac{1}{n}\sum_{i\in\mathbb{V}} \bm{h}_i^{(L)}, $$
which we referred to as the {\itshape Readout} step in Section~\ref{sec:model}. 
$\bm{g}$ or $\bm{H}^{(L)}$ is then fed into a fully-connected network (MLP) to be transformed into the desired form of output for further assessment, e.g., a scalar value as a prediction score in graph regression. We denote this network as $f(\cdot)$, which, in our experiments, is fixed to be a three-layer MLP of the form
\begin{displaymath}
f(x) = \sigma(\sigma(x \bm{W}_1)\bm{W}_2)\bm{W}_3, 
\end{displaymath}
where $\bm{W}_1\in\mathbb{R}^{p \times (p/2)}$, $\bm{W}_2\in\mathbb{R}^{(p/2) \times (p/4)}$, $\bm{W}_3\in\mathbb{R}^{(p/4) \times k}$ with $k$ being the desired output dimension. 
The final output, either $f(\bm{g})$ or $f(\bm{H}^{(L)})$, is compared to the ground-truth by a task-specific loss function. For graph classification and node classification, we choose cross-entropy loss and for graph regression, we use mean absolute error.   % (or the $L1$ loss)

% These results are further supported by the findings on the GIN architecture displayed in Table \ref{tab:gin-nclass}.

\end{document}